\documentclass[sigconf]{acmart}

\renewcommand\footnotetextcopyrightpermission[1]{} % removes footnote with conference information in first column

\usepackage{booktabs} % For formal tables
\usepackage{listings}
\usepackage{hyperref}
\usepackage{float}
\usepackage{framed}
\usepackage{mdframed}
\usepackage{comment}
\usepackage{algorithm}
\usepackage{graphicx}
\usepackage{amsmath}
\usepackage{amsfonts}
\usepackage{slashbox}

\begin{document}

\newcommand{\para}[1]{\textbf{#1:}}

% Copyright
%\setcopyright{acmcopyright}
%\setcopyright{acmlicensed}
%\setcopyright{rightsretained}
%\setcopyright{usgov}
%\setcopyright{usgovmixed}
%\setcopyright{cagov}
%\setcopyright{cagovmixed}

% DOI
%\doi{10.475/123_4}

% ISBN
%\isbn{123-4567-24-567/08/06}

%Conference
%\conferenceinfo{CoDS-COMAD '18}{Jan 11--13, 2018, Goa, India}

%\acmPrice{\$15.00}

%
% --- Author Metadata here ---
%\conferenceinfo{WOODSTOCK}{'97 El Paso, Texas USA}
%\CopyrightYear{2007} % Allows default copyright year (20XX) to be over-ridden - IF NEED BE.
%\crdata{0-12345-67-8/90/01}  % Allows default copyright data (0-89791-88-6/97/05) to be over-ridden - IF NEED BE.
% --- End of Author Metadata ---

%\title{Alternate {\ttlit ACM} SIG Proceedings Paper in LaTeX
%format\titlenote{(Produces the permission block, and
%copyright information). For use with
%SIG-ALTERNATE.CLS. Supported by ACM.}}
%\subtitle{[Extended Abstract]
%\titlenote{A full version of this paper is available as
%\textit{Author's Guide to Preparing ACM SIG Proceedings Using
%\LaTeX$2_\epsilon$\ and BibTeX} at
%\texttt{www.acm.org/eaddress.htm}}}
%

\setcopyright{acmcopyright} % if you give the rights to ACM
\acmDOI{...} % DOI - Insert your DOI below...
\acmISBN{...} % ISBN - Insert your conference/workshop's ISBN below...
\acmYear{2018} % Insert Publication year
\copyrightyear{2018} % Insert Copyright year (typically the same as above)
\acmPrice{15.00}
\acmConference[CoDS-COMAD]{}{Jan 11-13, 2018}{Goa, India}

\title{Fault in your stars: An Analysis of Android App Reviews}
%\title{Mismatch Detection in Mobile App Reviews using Dependency based Convolutional Neural Networks}

\author{Rahul Aralikatte}
\affiliation{\institution{IBM Research}}
\email{rahul.a.r@in.ibm.com}

\author{Giriprasad Sridhara}
\affiliation{\institution{IBM Research}}
\email{girisrid@in.ibm.com}

\author{Neelamadhav Gantayat}
\affiliation{\institution{IBM Research}}
\email{neelamadhav@in.ibm.com}

\author{Senthil Mani}
\affiliation{\institution{IBM Research}}
\email{sentmani@in.ibm.com}

\begin{abstract}
Mobile app distribution platforms such as Google Play Store allow users to share their feedback about downloaded apps in the form of a review comment and a corresponding star rating. Typically, the star rating ranges from one to five stars, with one star denoting a high sense of dissatisfaction with the app and five stars denoting a high sense of satisfaction.

Unfortunately, due to a variety of reasons, often the star rating provided by a user is inconsistent with the opinion expressed in the review. 
For example, consider the following review for the Facebook App on Android; ``Awesome App". One would reasonably expect the rating for this review to be five stars, but the actual rating is one star! 

Such inconsistent ratings can lead to a deflated (or inflated) overall average rating of an app which can affect user downloads, as typically users look at the average star ratings while making a decision on downloading an app. Also, the app developers receive a biased feedback about the application that does not represent ground reality. This is especially significant for small apps with a few thousand downloads as even a small number of mismatched reviews can bring down the average rating drastically.

In this paper, we conducted a study on this review-rating mismatch problem. We manually examined 8600 reviews from 10 popular Android apps and found that 20\% of the ratings in our dataset were inconsistent with the review. Further, we developed three systems; two of which were based on traditional machine learning and one on deep learning to automatically identify reviews whose rating did not match with the opinion expressed in the review. Our deep learning system performed the best and had an accuracy of 92\% in identifying the correct star rating to be associated with a given review.

In another evaluation, we asked 23 end users to write reviews for any 5 apps that they had used recently. We got 115 reviews from 66 different mobile apps. Our deep learning system had an accuracy of 87\%. 

Further, our study suggests that this problem is quite prevalent among apps. Across the ten apps used in our study, the mismatch percentage ranged from 16\% to 26\%.
\end{abstract}

\keywords{Android; Mobile apps; Convolutional Neural Networks; Machine Learning; Deep Learning}

\maketitle

\section{Introduction}
\label{sec:intro}
%introduction.tex

\begin{comment}
As smart phones become ubiquitous, more and more apps are being developed for these phones. It is estimated that the mobile handset industry is growing at greater than 20\% per year~\cite{VisionMobile1}. iOS and Android are very lucrative software platforms, with average monthly revenue of nearly \$5,000 and nearly \$3,500 per app, respectively. 
Additionally, the developers' mindshare index shows that iOS and Android are the top two software platforms being used by developers worldwide~\cite{VisionMobile1,VisionMobile2}. There are about a million apps apiece for Android and the iOS versions.

Mobile apps are typically available for download at digital distribution platforms like Google Play Store\footnote {\url {https://play.google.com/store?hl=en}} and Apple Store. These platforms allow a user to not only download apps but also make informed choices about downloading apps. They allow users to search for apps and display various details about an app such as the total downloads of the app so far, related apps, description, screenshots, etc.
\end{comment}

Mobile apps are typically available for download at digital distribution platforms like Google Play Store and Apple Store. Once a user has downloaded and used an app, these %digital
distribution platforms %such as Google Playstore and Apple Store 
also allow the user to enter feedback about the app. The feedback is received in the form of a review comment and an associated star rating. The star rating ranges from one to five stars with one star denoting extreme dissatisfaction with an app and five stars denoting high satisfaction.

The review comments and star ratings are very important as studies and our survey show that users typically download an app based on these factors~\cite{appStoreMiningMsr2012}. 
%Users make decisions about downloading an app based on various factors such as their requirement, whether the app is free or paid, number of downloads of the app so far, description and screenshots, %of the app, size of the app, permissions required and so on. However, according to studies~\cite{appStoreMiningMsr2012}, one of the most important factors in downloading an app is the rating of the app, which is computed as an average of the individual star ratings provided by prior users of the app.

As ratings are an important factor in determining the download of an app, it is imperative that ratings be accurate i.e., a rating accurately reflects the experience of the user with the app. 

However, a study~\cite{Dave:2003:MPG} and our investigations suggest that often the star rating given by a user is not consistent with the opinion expressed in the review comment. Consider the following review text for the Instagram app on Android.
\begin{framed}
\textit{ ``Love instagram it's the best in the world Love it it's the best in the world''} (sic)
\end{framed}

One would reasonably expect that due to the highly positive sentiment expressed in the review, the associated star rating would be five stars, but the actual star rating is one star! Such mismatches bring down the average rating of an app, which can adversely affect future downloads of the app (especially for small and upcoming apps without many downloads).

\begin{comment}
We can also have reviews such as the following
\begin{framed}
\textit{``Notifications? Notifications stopped working. The only one I get is for primary or priority inbox. No longer able to use the sorting feature and get notifications for each group. Doesn't allow turning on label notifications. Pop up  appears,  but does not work.". } (sic)
\end{framed}

This lengthy review is full of complaints with one complaint following another. One would expect that such a review would be accompanied by a one star rating, but the actual rating provided by the user is five stars! Such reviews would have the effect of artificially increasing the average rating of an app. If we have many such reviews, the average rating would be way higher than deserved and in the long run, users would no longer trust the average rating of the app at all.
\end{comment}

Review rating mismatches can occur due to a variety of reasons; one reason could be that novice end users may simply be confused about the difference between one and five stars~\cite{Dave:2003:MPG}. 
%Thus, they may write a positive review but leave behind a rating of one star as they think that one star is better than five stars~\cite{Dave:2003:MPG}. 
On the other end, a negative opinion accompanied by five stars could happen due to the following reason: A user may initially provide a rating of five stars to an app due to a positive experience. Review systems allow users to simply rate without an explicit review comment. Later on, the user may have a negative experience with the app, (usually after an update). He may then write about his problems with the app, but may \textit{forget} to update the rating to accurately reflect his current review text. Thus, a review with a largely negative opinion can have a high rating of five stars. This hypothesis is in fact confirmed by our survey responses in Section \ref{sec:motiv}

In this paper, we perform a study of this review rating mismatch problem. We first establish by means of a user survey and manual study that the review rating mismatch problem is prevalent across popular apps on Android. We also establish the need for a system which can \textit{automatically} detect inconsistencies between reviews and ratings. We then show that the development of such an automated system is non-trivial i.e., simple techniques such as natural language sentiment analysis do not suffice.

We then empirically establish that our system performs well i.e., can accurately identify reviews whose ratings do not match with the opinion expressed in the review text. 
%Validation is performed both through cross validation and an external user evaluation process. 

We use our automated system to find the prevalence of review-rating mismatches across 10 popular apps on Android and discover that 16\% to 26\% of the ratings do not match with their reviews. We finally show the generalizability of our system by analyzing mismatches on datasets of completely different domains.

To summarize, the main contributions of this paper are as follows:
\begin{itemize}
\itemsep0em
\item A survey of Android app end users and developers which suggests that:
\begin{itemize}
\item Review text and star rating should match
\item It is useful to have an automated system to detect mismatches
\item Users do not update their rating when they change the review text
\end{itemize}
\item A manual investigation of 8600 reviews from 10 popular apps on Android. This study shows that about 20\% of the reviews have inconsistent ratings and this inconsistency is distributed across apps.
\item Machine learning and deep learning techniques to automatically detect mismatched review-ratings.
\item A deep learning model which achieves a cross-validation accuracy of 92\% in identifying reviews with inconsistent ratings.
\item An evaluation with 23 independent human evaluators on a test set of 115 reviews drawn from 66 diverse mobile apps. The accuracy ranged from 84\% to 87\%
%\item An evaluation with 23 independent human evaluators on the accuracy of our systems, which show that our system achieves an accuracy of 74\% in identifying reviews with inconsistent ratings (on a test set of 115 reviews drawn from 66 mobile apps)
\item An estimate of the prevalence of review-rating mismatches across 10 popular Android apps using our deep learning system. The mismatch ranges from 16\% to 26\%
%\item Across 10 popular apps on Android, an estimate of the number of reviews whose rating is not consistent with the opinion expressed in the review. %and present our results and recommendations. 
%This shows that the problem is widely prevalent, for instance, in Facebook, about 19\% of the reviews have mismatched ratings. Across the ten apps used in our study, the mismatch percentage ranged from 16\% to 26\%.
\end{itemize}

%Our online tool to predict review rating is available at\\ http://mismatch.mybluemix.net/. 

The remainder of this paper is organized as follows: Section \ref{sec:motiv} provides the motivation, Section \ref{sec:approach} describes our approach to solving this review rating mismatch problem, Section \ref{sec:eval} describes the evaluation of our approach.
Section \ref{sec:impl} discusses the implications of our work
and Section \ref{sec:relatedwork} describes the related work (which has mainly focused on extracting feature requests and bugs from reviews and not on detecting inconsistent review-ratings). 
%Finally, Section \ref{sec:conc} describes the conclusion.
We conclude in Section \ref{sec:conc}.

%We believe that the whole notion of average rating needs refinement, one refinement proposed in this paper is to detect inconsistencies between review text and rating, however, other refinements are possible and can be the subject for future work which is described in Section \ref{sec:future}. 

%
\section{Motivation}
\label{sec:motiv}
%motivation.tex

In this section, we provide motivation for our work using two methods:
\begin{itemize}
\item A survey of Android app end users and developers
\item A manual annotation of reviews from popular Android apps
\end{itemize}

\subsection{Motivating Survey}
In our survey, we primarily wanted to know whether users believed that a mobile app star rating and associated review text should match, whether an automated system to detect mismatched reviews is useful and whether users update the star rating when they update the review text.

We hosted the survey questions on Google Forms and posted the link on different platforms such as Android forums, mailing lists, bulletin boards of the Computer Science Department at two premier
universities in our country and organization. %India and our research lab. 

No compensation was provided to any of the survey participants. They were not told about our hypothesis about review-rating mismatch. The survey had two branches based on whether the respondent was an Android app developer or only an end user. The end users had seven questions while the developers had four questions. 

We received 109 responses to our survey with 82\% being end users and 18\% developers. The survey responses are shown in Tables~\ref{table:survey-q-rating-review-match},~\ref{table:survey-q-utility-of-automated-system},~\ref{table:survey-end-users} and~\ref{table:survey-developers}.

\textit{The fundamental premise of our work that the star rating and the associated review text should correspond is strongly supported by the responses} shown in Table~\ref{table:survey-q-rating-review-match}. Further, Table~\ref{table:survey-q-utility-of-automated-system} suggests that both end-users and developers feel that an automated system to detect mismatched review-ratings is useful.

Table~\ref{table:survey-end-users} shows the other survey questions to end users and their responses. It suggests that users base their download decision on existing reviews and average rating. Thus, if we have a number of ratings that are inconsistent with the review text
(say, the correct rating should have been five, but the user rated as one),
the average rating of the app may decrease which will in turn will affect app downloads. For widely popular apps with millions of downloads, it is possible that inconsistent ratings do not affect
the average rating in a significant manner; however, for small and upcoming apps, the average rating will be affected leading to decreased downloads.

The last row in  Table~\ref{table:survey-end-users} is very instructive. It suggests that users typically do not update the rating after updating their review text. We believe this is one of the major causes of review-rating mismatch.

Table~\ref{table:survey-developers} shows the other survey questions to app developers and their responses. It suggests that developers believe that review-rating mismatch is prevalent and importantly, affects app development.

To conclude, both Android app end users and developers agree that the review text and associated star rating must match. Further, they consider an automated system to detect mismatched reviews-ratings is useful.

One issue with surveys is that \lq\lq{}what people say\rq\rq{} could be different from \lq\lq{}what people do\rq\rq{}~\cite{selecting-empirical-methods-for-se-research}. To overcome such issues, typically, a \emph{triangulation} approach is used to confirm a survey\rq{}s findings~\cite{selecting-empirical-methods-for-se-research}. Thus, we also manually annotated review text from popular Android apps and rated them to confirm that the review-rating mismatch problem is indeed prevalent. We describe our procedure in the next sub-section.
%expressed in the survey.
%We also found that in the projects that the survey takers were responsible for, there were about 2000 task comments, which further supports the opinions
%expressed in the survey.

\begin{table}[htbp]
\scriptsize
\begin{tabular}{ |l || l | l |}
\hline
\textbf{Should the star rating and review text match? } & \textbf{Yes (\%) } & \textbf{No (\%)}  \\ \hline \hline
Opinion of End Users  & \textbf{92} & 8 \\ \hline
Opinion of Developers & \textbf{79} & 21  \\ \hline
\end{tabular}
  \caption{Survey: Should the star rating and review text match?}
    \label{table:survey-q-rating-review-match}
\end{table}

\begin{table}[htbp]
\scriptsize
\begin{tabular}{ |l || l | l |}
\hline
\textbf{Utility of an automated system to detect} & \textbf{End Users} & \textbf{Developers}  \\
\textbf{star rating and review text mismatch} & & \\ \hline \hline
Very Useful  & \textbf{26} & \textbf{42} \\ \hline
Somewhat Useful  & \textbf{32} & \textbf{31} \\ \hline
Marginally Useful  & 28 & 6 \\ \hline
Not Useful  & 14 & 21 \\ \hline
\end{tabular}
  \caption{Survey: Utility of an automated system to detect inconsistent ratings (Results in Percentages)}
    \label{table:survey-q-utility-of-automated-system}
\vspace{-10pt}
\end{table}

\begin{comment}

	\begin{table*}[htbp]
	\scriptsize
	\begin{tabular}{ |l || l | l | l | l |}
	\hline
	%Question & Always & Most of the time & Occasionally & Never  \\ \hline \hline
	\textbf{Question} & \textbf{Always} & \textbf{Most times} & \textbf{Occass-} & \textbf{Never}  \\ 
	 & & & \textbf{ionally} &	\\ \hline \hline
	Do you download an app based on existing reviews? & 15 & 41 & 37 & 7 \\ \hline
	Does the average rating of an app influence your decision to download an app? & & & \\ \hline
	Do you rate apps on App/Play Store?& 37 & 22 & 37 & 4   \\ \hline
	Do you also write review comments along with rating?  & 4 & 4 & 33 & 59   \\ \hline
	If you update your rating for an app,do you update your review? & & & &  \\
	Is it useful to have a system that analyzes a review and & & & &  \\
	suggests the appropriate rating to be given?  & & & &  \\ \hline
	\hline
	\end{tabular}
	  \caption{Survey on Task Comments with Experienced IBM Developers : Results in Percentages}
	    \label{table:survey-task-com-ibm}
	\end{table*}

\end{comment}

%\begin{comment}

\begin{table}[htbp]
\scriptsize
\begin{tabular}{ |l || l | l | l | l |}
\hline
%Question & Always & Most of the time & Occasionally & Never  \\ \hline \hline
\textbf{Question} & \textbf{Always} & \textbf{Mostly} & \textbf{Some-} & \textbf{Never}  \\ 
 & & & \textbf{times} &	\\ \hline \hline
Do you download an app based & 15 & \textbf{56}  & 25 & 4 \\ 
on existing reviews? & & & & \\ \hline
Does average rating of an app & 23 & \textbf{56} & 16 & 5  \\
influence your decision & & & &  \\
to download the app? &  &  &  &    \\ \hline
Do you rate apps on App/Play Store? & 2 & 21 & \textbf{58}  & 19   \\ \hline
Do you write review comments along  & 5 & 5 & \textbf{47}  &  41  \\ 
with rating? & & & & \\ \hline
If you update your review for an app, & & & &  \\
do you update your rating? & 11  & 8  & 16 &  \textbf{65}  \\ \hline
\end{tabular}
  \caption{Survey with mobile app end users (Results in Percentages)}
    \label{table:survey-end-users}
\vspace{-10pt}
\end{table}

%\end{comment}

\begin{table}[htbp]
\scriptsize
\begin{tabular}{ |l || l | l | l |}
\hline
%Question & Always & Most of the time & Occasionally & Never  \\ \hline \hline
\textbf{Question} & \textbf{Yes} & \textbf{Somewhat} &  \textbf{No}  \\ \hline \hline
Is review-rating mismatch prevalent?& \textbf{42} & \textbf{42} & 16  \\ 
Does review-rating mismatch affect app development & \textbf{37} & \textbf{53} & 10  \\
%affect app development & & &  \\
\hline
\end{tabular}
  \caption{Survey with mobile app developers (Results in Percentages)}
    \label{table:survey-developers}
\vspace{-10pt}
\end{table}

%\subsection{Data Analysis}
\subsection{Manual Annotation of Reviews}
\label{label-man-ann}

%Studies such as the seminal work done in~\cite{Dave:2003:MPG} have suggested that there tends to be a mismatch between reviews and rating. It suggests
%that the review rating mismatch happens due to novice users being
%confused about the star rating system and thinking that one star is better than five stars. As described in the Introduction, we believe
%that review rating mismatch also happens due to an update of the review comment by an user without updating the previously assigned star rating.
%
%However, we wanted to be certain that the review rating mismatch problem happened often enough to warrant an automated solution i.e., we were concerned
%if the mismatches we saw were stray incidents and thus anomalies. To satisfactorily answer this question, we decided to manually annotate reviews from a large number of apps. 

We chose 8600 random reviews from 10 very popular Android apps. Some characteristics of the apps are shown in Table \ref{table:apps}.
As can be seen from the table, we have diversity in the sample with apps drawn from different categories such as social media, e-mail, games and so on.

\begin{table}
\scriptsize
\centering
\begin{tabular}{ |c|c|c|c| } 
 \hline
 App & Avg. rating & \# downloads & \# Reviews \\ 
 \hline \hline
 Facebook & 4 & 1,000,000,000 & 40,632,882 \\ 
 Gmail &  4.3 & 1,000,000,000 &  2,485,656 \\ 
 Google Plus & 4.2 & 1,000,000,000 & 2,458,398 \\   
 Twitter & 4.2  & 500,000,000 & 7,345,190 \\ 
 Subway Surfers & 4.4  & 500,000,000 & 19,025,802 \\ 
 Instagram & 4.5 & 500,000,000 & 31,922,875 \\ 
 Angry Birds & 4.4 & 100,000,000 & 4,748,194\\   
 %ClashOfClans & 4.6 & 100,000,000 &  25,342,416 \\ 
 Temple Run & 4.3 & 100,000,000 & 2,970,608 \\ 
 LinkedIn & 4.2 &50,000,000 & 869,962\\ 
 Quora & 4.5  & 1,000,000&  135,060\\ 
 \hline
\end{tabular}
\caption{Popular Android Apps on Google Play Store used in our Study}
\label{table:apps}
\vspace{-15pt}
\end{table}

Once we chose the reviews, three annotators set about manually annotating them. The annotation task is to read the review and assign a star rating ranging from one to five stars, without having seen the original star rating. Assigning a star rating is a somewhat subjective process; hence we formulated the following guidelines about star ratings corresponding to review text.

\begin{itemize}
\item \textbf{Five Stars:} A five star rating is assigned to review text which are entirely complimentary without any reports of a problem or even a feature request. For example, the Facebook review, 

%\mbox{\textit{ ``Awesome!''}}
\begin{framed}
\textit{ ``Awesome!''}
\end{framed}

\item \textbf{Four stars:} A four star rating is assigned to reviews which are almost like five star rated reviews but which express a feature request (through words such as `wish'). For example, the Gmail review,

\begin{framed}
\textit{ ``Great app! Wish we could delete all mails with one click''}
\end{framed}

\item \textbf{Three stars:} A three star rating is given to those reviews which are in between praising and criticizing. For example, the Temple Run review,

\begin{framed}
\textit{ ``Nice app but annoying with too many ads''}
\end{framed}

\item \textbf{Two stars:} A two star rating is given to those reviews which in general are not complimentary but do not sound extremely dissatisfied i.e., find some redeeming features as well. For example, the Quora review, 

\begin{framed}
\textit{ ``Ok, but very slow and crashes at times"}
\end{framed}

\item \textbf{One star:} A one star rating is given to those reviews which express extreme disappointment with an app and finds nothing redeeming. For example, the Temple Run review, 

\begin{framed}
\textit{ ``Hate this app! Uninstalling now"}
\end{framed}

\end{itemize}

Armed with the above guidelines, we set about manually annotating reviews. Note that this is a tedious process and the annotators could annotate only about 3 reviews per minute. \textit{This time consumption further illustrates the need for an automated solution.} 

We measured the agreement among the annotators through 
%Cohen's Kappa Test (for each pair of annotators) and 
the Fleiss's Kappa, a standard inter-annotator agreeent measure 
when there are multiple annotators~\cite{CJS:CJS3}. The Kappa score
was 0.7 indicating a \textit{substantial} level of agreement~\cite{CJS:CJS3}. 

For reviews where annotators were not in agreement, we took the majority rating if at least two annotators agreed, 
else we took the average as the final rating.

%Table~\ref{table:ratings-dist} shows the original distribution of the 8600 %reviews among the ratings 1 to 5 (as assigned by the original users) (third %column) and the distribution according to annotators (fourth column). 

Table~\ref{table:ratings-mismatch-distribution} shows the distribution of
the original rating and the annotator rating. The diagonal elements represent agreement between the original reviewer and the annotators while the
non-diagonal elements represent the disagreements.

\begin{table}[t]
\centering
\small
\begin{tabular}{ |c|c|c|c|c|c||c| } 
\hline 
% & 1 & 2 & 3 & 4 & 5 \\ \hline
\backslashbox{Ann}{Org}
%&\makebox[3em]{1}&\makebox[3em]{2}&\makebox[3em]{3} &\makebox[3em]{4}&%%%\makebox[3em]{5}\\ \hline\hline  \\
  & 1 & 2 & 3 & 4 & 5 & Total\\ \hline \hline
1 & 1337 & 367  & \textbf{238}  & \textbf{84}  & \textbf{73}  & 2099 \\
2 & 314 & 313 & \textbf{257}  & \textbf{168}  & \textbf{89}  & 1141 \\
3 &  \textbf{74} & \textbf{57} & 397 & \textbf{275}  & \textbf{229}  & 1032\\ 
4 & \textbf{15} & \textbf{5}  & \textbf{52}  & 434 & 510  & 1016 \\
5 &  \textbf{38} & \textbf{15} & \textbf{71} & 438 & 2750 & 3312 \\ \hline
Total &  1778 & 757 & 1015 & 1399 & 3651 & 8600 \\
 \hline
\end{tabular}
\caption{Manually annotated reviews: Distibution of Org (Original Rating) vs. Ann (Annotator Rating)}
\label{table:ratings-mismatch-distribution}
\vspace{-10pt}
\end{table}

%\begin{table}
%\scriptsize
%\centering
%\begin{tabular}{ |c|c|c|c| } 
% \hline
% Type & Rating & Original Count & Annotation count \\ 
% \hline \hline
% Bad & 1 & 1776 & 2103 \\ 
% & 2 &  755 & 1140 \\ 
% \hline
% Neutral  & 3 &  1015 & 1031 \\ 
% \hline
%Good  & 4 &  1398 & 1016 \\   
% & 5 &  3656 & 3310 \\ 
% \hline
%\end{tabular}
%\caption{Distribution of star ratings in manually annotated review set (8600 reviews)}
%\label{table:ratings-dist}
%\vspace{-20pt}
%\end{table}

We then proceeded to count the mismatches between the original rating and the rating assigned by the annotators. 
%In calculating this mismatch, we made a reasonable decision that if there was a difference of \emph{only one} between the original rating and annotator's rating, we would \emph{not} count it as a mismatch. Thus, for example, if the user gave a rating of 5, but annotator rating was 4, we do not count it as a mismatch. Such a scheme ensures that fine differences in opinion are tolerated and we do not get an inflated sense of mismatches. We considered mismatches only where the difference was greater than one (for example, original rating was 5, while annotator rated 2) and 
We classified ratings 5-4 as \textit {Good}, 3 as \textit{Neutral} and 1-2 as \textit{Bad}. \textit{We counted the mismatches only when the rating for a review text moved from one category to another}. For example, if the original rating was 5 and the annotator rating was 3, we considered it as a mismatch. However, if the annotated rating was 4, we did not consider it as a mismatch. 

Such a scheme ensures that fine differences in opinion are tolerated and \textit{we do not get an inflated sense of mismatches}.

Using the above methodology, we found about 20\% mismatches in the 8600 reviews. The highlighted cells in Table~\ref{table:ratings-mismatch-distribution} show the mismatch count. The mismatch percentage per app are shown in Table~\ref{table:ratings-mismatch-dist} indicating that the review-rating mismatch problem is fairly prevalent and also occurs across different apps. A sample of review-rating mismatches are shown in Table~\ref{table:ratings-mismatch-sample}.

To conclude, the survey of end-users and app developers suggested that the problem of review text and star rating mismatch was fairly prevalent. Our manual annotation of reviews across popular apps also suggest the same. 

We also found during the manual annotation of review text, identifying mismatches is a tedious process which supports the survey findings of the need for an automated solution. Thus, the manual annotation has reinforced the survey findings.

\begin{comment}
\begin{table}
\scriptsize
\centering
\begin{tabular}{ |c|c| } 
 \hline
App & \% mismatch \\ 
\hline
Angry Birds & 21.77 \\ 
Facebook  & 12.9 \\ 
Gmail & 15.79 \\  
Google Plus  & 14.12 \\ 
Instagram & 12.72 \\ 
LinkedIn & 10.92\\ 
Quora & 9.5\\ 
Subway Surfers & 10.3 \\ 
Temple Run & 12.35 \\ 
Twitter & 12.74 \\   
 \hline
\end{tabular}
\caption{Distribution of mismatches per app in manually annotated review set}
\label{table:ratings-mismatch-dist}
\end{table}
\end{comment}

\begin{table}
\scriptsize
\centering
\begin{tabular}{ |c|c||c|c| } 
 \hline
App & \% mismatch & App & \% mismatch \\ 
\hline \hline
Angry Birds & 20.04 & LinkedIn & 18.43\\ 
Facebook  & 21.36 & Quora & 19.81\\
Gmail & 24.33 &  Subway Surfers & 16.31 \\ 
Google Plus  & 19.85 & Temple Run & 20.5 \\ 
Instagram & 18.94 & Twitter & 21.79 \\   
\hline
\end{tabular}
\caption{Distribution of mismatches per app in manually annotated review set}
\label{table:ratings-mismatch-dist}
\vspace{-10pt}
\end{table}

%\begin{table}[t]
%\scriptsize
%\centering
%\begin{tabular}{ |c|c|c| } 
% \hline
% Review & User  & Annotator  \\ 
%   & Rating & Rating \\
% \hline \hline
% I like this game, its really & & \\
%  awesome!  & 1 & 5 \\ 
% \hline
% Love instagram it's the & & \\
%  best in the world & 1 & 5 \\
% \hline
% Nice one.  One of & & \\
%  the best informatory app & 1 & 5 \\
%  \hline
% Crashing all the time & 5 & 1 \\
% \hline
% AHHH  IT WONT LET ME  & & \\
% CHANGE MY PROFILE & & \\
% PICTURE ANYMORE!!!!!!!!!!!!! & 5 & 1 \\
% \hline
% New update really sucks......very bad  & 5 & 1 \\
% \hline
% I love instagram but i & & \\
%  cannot post video & 5 & 3 \\
%  \hline
% Nice Enjoying & 2 & 5 \\
% \hline
% Notification?  Keeps telling & & \\
% me I have a notification & & \\
%  when I don't. & 5 & 2 \\
%  \hline
% App crashes in Marshmallow.  & 4 & 1 \\
% \hline
%\end{tabular}
%\caption{Sample of mismatched reviews in manually annotated review set}
%\label{table:ratings-mismatch-sample}
%\vspace{-15pt}
%\end{table}

\begin{table}[t]
\scriptsize
\centering
\begin{tabular}{ |c|c|c| } 
 \hline
 Review & User  & Annotator  \\ 
   & Rating & Rating \\
 \hline \hline
 %I like this game, its really & & \\
  %awesome!  & 1 & 5 \\ 
% \hline
 Love instagram it's the & & \\
  best in the world & 1 & 5 \\
 \hline
 %Nice one.  One of & & \\
  %the best informatory app & 1 & 5 \\
  %\hline
 Crashing all the time & 5 & 1 \\
 \hline
 AHHH  IT WONT LET ME  & & \\
 CHANGE MY PROFILE & & \\
 PICTURE ANYMORE!!!!!!!!!!!!! & 5 & 1 \\
 \hline
 New update really sucks......very bad  & 5 & 1 \\
 \hline
 I love instagram but i & & \\
  cannot post video & 5 & 3 \\
  \hline
 Nice Enjoying & 2 & 5 \\
 \hline
 Notification?  Keeps telling & & \\
 me I have a notification & & \\
  when I don't. & 5 & 2 \\
  \hline
 App crashes in Marshmallow.  & 4 & 1 \\
 \hline
This game is very nice but it hangs sometimes.. & 2 &3 \\
 \hline
really love this app but severe problem of battery drain & 4 & 3 \\
 \hline
 angry birds  nice game  & 	3 & 5 \\
 \hline
\end{tabular}
\caption{Sample of mismatched reviews in manually annotated review set}
\label{table:ratings-mismatch-sample}
\vspace{-15pt}
\end{table}

%Now that we have empirically established that the problem of review rating mismatch is fairly prevalent and %manually identifying such mismatches is a tedious process, we describe our automated solution in the
%next section.

%Now that we have empirically established that the problem of review rating mismatch is fairly prevalent and %manually identifying such mismatches is a tedious process, we list the various machine learning approaches %that can be used for building an automated solution.  However, to motivate the need for a sophisticated %machine learning approach, we first explored an automated solution using sentiment analysis. 

%To motivate the need for sophisticated learning approaches, we first explored an automated solution using %natural language sentiment analysis. 

Moving on, a simple or naive approach for automation is to use natural language sentiment analysis. We describe in the next subsection as to why this approach does not suffice and hence motivate the need for
sophisticated automated solutions.

%\subsubsection{Why Sentiment Analysis Based Rating Prediction is insufficient?}
\subsection{Sentiment Analysis Based Rating Prediction is insufficient}

In Natural Language Processing, sentiment analysis research deals with automatically analyzing the sentiment expressed in a sentence. Usually, the sentence analyzed is categorized into one of 5 categories viz., \textit{highly negative}, \textit{negative}, \textit{neutral}, \textit{positive}, and \textit{highly positive}. 

Intuitively, it appears that one can thus apply sentiment analysis to review text, obtain a category such as \textit{highly positive} and map it to a numerical star rating, \emph{5}. However, this approach
does not work in practice as described below.

{\bf Sentiment Score Calculation:} For each review in our set of 8600 reviews, we first applied a natural language tokenizer from the Stanford NLP toolkit~\cite{manning-EtAl:2014:P14-5} to obtain individual sentences. We then applied sentiment analysis to each sentence and computed the overall average sentiment as follows: The five sentiment categories were mapped to an ordinal scale ranging from 1 to 5, with the category \textit{highly negative} mapped to 1 and the category \textit{highly positive} mapped to 5. Let $s_i$ be the sentiment score for the $i^{th}$ sentence of the review text. Then the average sentiment score for the entire review text, $\textbf{S}$, is given by

\begin{equation}
\textbf{S} = round \left (\frac{1}{n} \sum_{i}{s_i} \right)
\end{equation} 

%We added the numerical sentiment score value obtained for each sentence and divided this sum by the total number of sentences in the review to obtain an average sentiment score, which was rounded to the nearest integer.
%which then was mapped back to the nominal scale. 

%, which in turn would be mapped back to the category, \textit{highly positive}.

For each review, we then found the correlation between the rating and $\textbf{S}$. We used both the Pearson and Spearman correlation. The Pearson correlation is a measure of the linear correlation between two variables $X$ and $Y$, giving a value between $+1$ and $-1$ inclusive, where $+1$ is total positive correlation, $0$ is no correlation, and $-1$ is total negative correlation. Similarly, the Spearman correlation assesses how well the relationship between two variables can be described using a monotonic function. If there are no repeated data values, a perfect Spearman correlation of $+1$ or $-1$ occur when each of the variables is a perfect monotone function of the other.

{\bf Results:} We obtained Pearson and Spearman correlation values of around $0.5$ each, for correlation between the ratings and the average sentiment scores. This indicates that there is some correlation (as expected) but not a very high degree, which \textit{precludes} the use of sentiment analysis alone to solve the problem of review-rating mismatch.

{\bf Discussion:} Our hypothesis as to why sentiment analysis does not work is as follows: Consider the review text for Facebook on Android, ``freezes after last update". There is nothing intrinsically negative about this sentence if one looks at it from a typical English sentence perspective. It is only in the domain of mobile apps that words like `freezes' have a very negative connotation. Sentiment analysis tools which are trained on standard English text will not be able to work accurately in this domain, thus preventing their use.

Since simple techniques like sentiment analysis do not suffice in automatically detecting review-rating mismatch, we explore more advanced solutions and describe them next. %in the next section.

%we list the various machine learning approaches that can be used for building an automated solution.  %However, to motivate the need for a sophisticated machine learning approach, we first explored an automated %solution using sentiment analysis. 

%
\section{Approach}
\label{sec:approach}
%approach

In this section, we describe our approach towards automatically detecting review-rating mismatches.
We use three different approaches, two of which are based on traditional machine learning and one which
is based on deep learning. The manually annotated set of 8600 reviews from 10 popular apps,
served as the training data for these approaches.

\subsection{Machine Learning}
\label{sec:ml}
Since the problem at hand is a five class classification problem, we first train different standard machine learning classifiers as baselines. We mention the different classifiers and the features used to train them below. 
%Machine learning algorithms use statistical methods to uncover patterns in data. These patterns are then used to either predict how future data may vary (regression) or for categorizing data into discrete buckets (classification). These algorithms need data to be converted into a suitable numerical format for processing. This conversion of data points into a list of numbers which represent their salient features is called \textit{feature engineering}. Predicting the star rating for a given review text is a typical classification problem. 

\subsubsection{Machine Learning algorithms used}
\begin{itemize}
\itemsep0em
\item Naive Bayes Classifier 
\item Decision Trees (J48)
\item Decision Stump (One-level decision tree)
\item Decision Table (Majority classifier)
\item AdaBoost (AdaBoost.M1, LogitBoost)
\item K-nearest neighbors (IBk)
\item Support Vector Machines (SMO)
\item Holte's 1R
\end{itemize}

%These classifiers have been used for number of tasks in software engineering. 
Due to space constraints, we do not provide details of these classifiers, but interested readers can refer to~\cite{Mitchell:1997:ML:541177} for more details.

\subsubsection{Features for machine learning}
\label{sec:features}

In supervised learning, along with manually annotated training data, we need to identify a proper set of features and extract feature values for each data point in the training data. As we are dealing with text data, it is natural to use TF-IDF scores as one of the features. But when we analyzed the 8600 reviews in our training data, there were some patterns which we felt would help represent the reviews better. We now describe the various features we extracted from our training set (apart from TF-IDF) to train the above mentioned classifiers and the intuition behind them:

\para{HasAllCapitalWords} When a user is unhappy, he tends to use all capital lettered words which is a norm on the world wide web. For example, consider the review from Facebook, ``NOTIFICATIONS STOPPED WORKING". The presence of such terms indicate frustration and disappointment and act as cue for a lower star rating. This is a binary $0$ or $1$ feature.

\para{HasNegativeCueWords} Similar to the above, this feature is also helpful in identifying reviews that should have a low star rating. Cue words such as \textit{crash, freeze, hang, slow, annoying,} etc. express a negative opinion. We manually constructed a dictionary of such negative cue words.

\para{HasQuestions} This feature is also helpful in identifying reviews that should have a low star rating. We observed that reviews which had a question typically indicated unhappiness and hence had lower star ratings. For example,`Why are there so many updates?'. Another way to identify such questions is to check for the presence of words like \textit{why, when, where, what,} etc. This is helpful in situations where users may have not used a question mark in the review text.

\para{HasExclamation} In contrast with the above features which deal with cues about identifying low star ratings, this feature deals with identifying high star ratings. We noticed that review text which were \textit{correctly} rated high often had exclamation(s). For example, the Facebook review,`awesome app!'.

\para{HasPositiveCueWords} This is a counterpart of
the \textit{HasNegativeCueWords} and is helpful in identifying reviews that should have a high star rating. Cue words that express a positive opinion such as \textit{great, excellent, awesome,} etc. are used here. We manually constructed a dictionary of such positive cue words.

\para{ReviewLength} We observed that reviews which are \textit{correctly} rated high tend to be short with only a few words. In contrast, reviews which have been \textit{correctly} rated low tend to be long (with many words) as the user typically is complaining about certain things. For example, \textit{`Notifications stopped working. The only one I get is for primary or priority inbox. No longer able to use the sorting feature and get notifications for each group. Doesn't allow turning on label notifications. Pop up  appears, but does not work'}. 

\para{SentimentScore} Although our empirical experiments suggest that sentiment analysis alone cannot be used to accurately predict star ratings, we believe, that in \textit{conjunction} with other features, it can help in accurately identifying review-rating mismatch. 

\para{ReadabilityScore} 
Our intuition is that, reviews that are \textit{correctly} rated high are more \textit{readable}~\cite{readability1} than those which are \emph{correctly} rated low. This is because, when users are unhappy or confronted with a problem, they may be agitated and hence may not write clearly. 
%\subsection{Model building and evaluation} 

\subsection{Deep Learning}
Usually feature engineering requires domain expertise which is often hard or expensive to come by. Also, when features are handcrafted, some important correlations may be missed which result in a poor representation of data which in turn decreases the accuracy of classification. This can be overcome by unsupervised feature learning/deep learning where the best feature representations are automatically learned from raw data.

RNNs with LSTM units~\cite{lstm} have become the defacto standard for unsupervised extraction of features from text. However, recently CNNs have been used to get state of the art results on problems involving small pieces of text~\cite{2016arXiv160906686R}. The app reviews which we are currently dealing with can often be short. Therefore we use a modified version of CNNs called Dependency based CNN (DCNN) for our classification problem. We briefly introduce CNNs below which is then followed by the internal working of the DCNN. 

\subsubsection{Convolutional Neural Networks} CNN~\cite{LeCun} is a type of feed-forward artificial neural network whose simplest form consists of 2 types of layers; the \textit{convolutional layer} and the \textit{pooling layer}. Neurons in each layer pass on their outputs to the next layer after they undergo a non-linear transformation (typically rectified linear or $tanh$). The neurons in a convolutional layer are connected to a small part of the adjacent layers which help in capturing spatially-local correlations. The pooling layer is used for non-linear down sampling of the inputs~\cite{DLTut1}. It also provides translation invariance; for example, it can identify a car, no matter in which way it is oriented.

\subsubsection{Dependency based Convolutional Neural Networks} 
Since CNNs were designed to operate mainly on images, they inherently apply convolution on continuous areas of inputs. If it is applied as is to language tasks, the convolution operates on the words in a sequential order. Let $x_i \in \mathbb{R}^d$ be a $d$-dimensional representation of a word (can either be a \textit{one-hot} representation or \textit{word2vec}~\cite{DBLP:journals/corr/abs-1301-3781}). If $\oplus$ is the concatenation operator, 

\begin{equation}
\textbf{x}_{ij} = \textbf{x}_i \oplus \textbf{x}_{i+1} \oplus ... \oplus \textbf{x}_{i+j}
\end{equation}

where $\textbf{x}_{ij}$ is the concatenated word vector from the $i^{th}$ to the $j^{th}$ word on which the convolution is applied. This is similar to n-gram models which feed local information to the convolution operations. In some cases, parts of a phrase maybe separated by several other words. Therefore we need a way to capture relationships among words even when they are not contiguous.

DCNN is similar to the model proposed in~\cite{DBLP:journals/corr/Kim14f}, but it also considers the ancestor and sibling words in a sentence's dependency tree. Here, we consider two types of convolutions: \textit{Convolutions of ancestor paths} and \textit{Convolutions on Siblings}.

\textit{Convolution of ancestor paths and siblings}: In this case, we concatenate word vectors as follows:

\begin{equation}
\textbf{x}_{ik} = \textbf{x}_i \oplus \textbf{x}_{p(i)} \oplus ... \oplus \textbf{x}_{p^k-1(i)}
\end{equation}

where $p^k(i)$ is a function which returns the $k^{th}$ ancestor of the $i^{th}$ word. Mathematically,

\begin{equation}
p^k(i) = \left\{\begin{matrix}
 p(p^{k-1}(i)) & \text{if} & k > 0 \\ 
 i & \text{if} & k = 0 
\end{matrix}\right.
\end{equation}

For a given $\textbf{x}_{ik}$, we apply a convolutional filter $\textbf{w} \in \mathbb{R}^{k\times d}$ with a bias term $b$.

\begin{equation}
c_i = \phi (\textbf{w} \cdot \textbf{x}_{ik} + b)
\end{equation}

where $\phi$ is a non-linearity such as $tanh$ or $ReLu$~\cite{DBLP:journals/corr/XuWCL15}. When this filter is applied on all the words in a sentence, we get a feature map $c \in \mathbb{R}^l$, where $l$ is the length of the sentence.

\begin{equation}
\textbf{c} = [c_1, c_2, \cdots, c_l]
\end{equation}

We repeat the same process for performing convolutions of siblings as well. The only difference is that, here $p^k(i)$ returns the $k^{th}$ sibling(s) rather than the $k^{th}$ ancestor.

\subsubsection{Our Model}: When using CNNs with text data, we use max-over-time pooling ~\cite{DBLP:journals/corr/Kim14f} to get the maximum activation over the feature map $c$. In DCNNs, we want the maximum activation from the feature map across the whole dependency tree (whole sentence). This is also called `max-over-tree' pooling~\cite{DBLP:journals/corr/MaHZX15}. Thus, each filter outputs only one feature vector after pooling. The final representation of a sentence will be many such features, one from each of the filters in the network. These feature vectors are finally passed on to a fully connected layer for classification.

The model we built is similar to what is done in~\cite{DBLP:journals/corr/MaHZX15}. We concatenated the feature maps obtained from ancestor path and sibling convolutions with the sequential n-gram representation. The concatenation with the sequential n-gram representation was done because the app reviews contains grammatical flaws which will result in parsing errors during dependency tree construction, but these parsing errors do not affect the sequential representation. The concatenated representation is shown in equation \ref{eqn:2}. Here $c_a^{(i)}$ represent the ancestor path feature map, $c_s^{(i)}$ represent the sibling feature map and $c^{(i)}$ represent the sequential feature map. We used 100 filters $(N_a = N_s = N = 100)$ for each representation.

\begin{equation}\label{eqn:2}
\textbf{c} = [\overbrace{c_a^{(1)}, \cdots, c_a^{(N_a)}}^\text{ancestors}; \overbrace{c_s^{(1)}, \cdots, c_s^{(N_s)}}^\text{siblings}; \overbrace{c^{(1)}, \cdots, c^{(N)}}^\text{sequential}]
\end{equation}

In our model, we used a dropout probability of 0.5 and a learning rate of 0.95 which was decayed using \textit{adadelta} update rule~\cite{DBLP:journals/corr/abs-1212-5701}. We used $100$ dimensional word embedding which were learned from scratch during training.

\section{Evaluation}
\label{sec:eval}
%evaluation

In this section, we describe our evaluation. We designed our evaluation to answer the following research questions (RQ):

\textbf{RQ1: Accuracy on manually annotated data:} What is the accuracy of our machine learning and deep learning techniques on manually annotated data?

\textbf{RQ2: Accuracy on data from the wild:} With what accuracy are we able to automatically predict the star rating of any given review text?

\textbf{RQ3: Mismatch Prevalence:} How prevalent are the review-rating mismatches across popular mobile apps for Android?

%\textbf{RQ4: Generalizability} How generalizable is our approach? What is the accuracy of our approach when applied on non mobile app reviews?

\begin{figure*}
\caption{RQ1: Comparison of accuracy measure between 9 different machine learning classifiers trained on handcrafted features and word vectors}
\label{figs:rq1}
\centering
\includegraphics[width=0.8\textwidth]{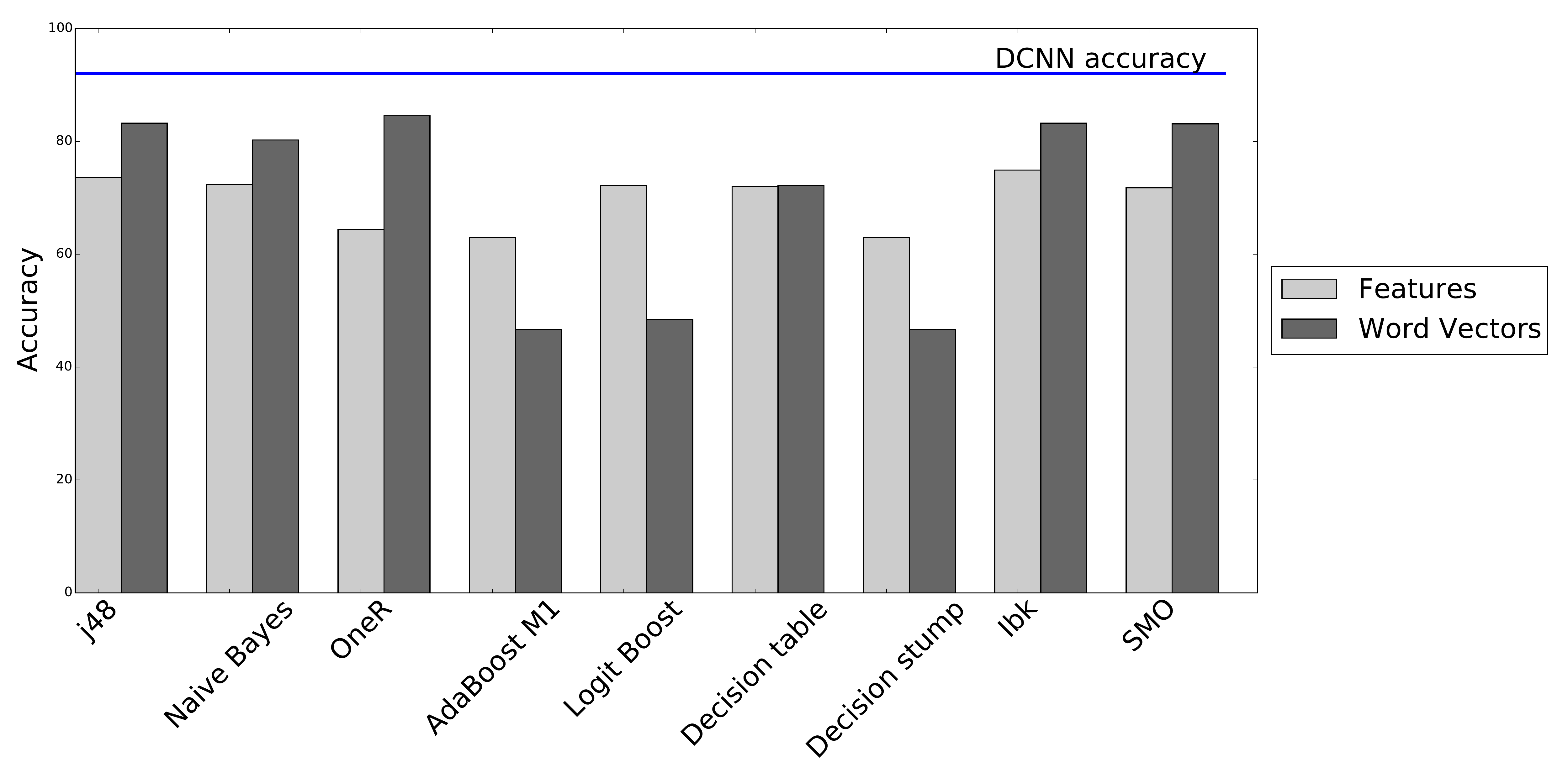}
\vspace{-10pt}
\end{figure*}

\subsection{Metric} 
%To compute \textit{accuracy}, we used the following approach: If the absolute difference of the predicted rating and the actual rating was less than one, then we consider the prediction as being accurate. Thus, for example, if the user gave a rating of 5, but our trained models predicted it as 4, then we do not count it as wrong prediction, and consider it as accurate. This helped to ensure that fine differences in opinion are tolerated and we do not get inflated sense of mismatches. 

%We considered \textit{mismatches} only when the difference was greater than one (for example user rated 5 while annotator rated 2 or 3). 
As explained in Section~\ref{label-man-ann} we categorized the ratings 5-4 as \textit{Good}, rating 3 as \textit{Neutral} and ratings 1-2 as \textit{Bad}. We considered our prediction to be accurate \textit{iff} the predicted rating and the correct rating fell in the same category. Therefore our accuracy is given by equation~\ref{eqn:acc} where $p_i$ is the predicted rating category, $c_i$ is the correct rating category and $n$ is the total number of reviews in the evaluation data.

Note that have not used "precision/recall" etc. as they make sense only when there are multiple correct answers and a subset of them is returned by an approach. Here we have only a binary answer as result (i.e., match/mismatch) and thus we strongly believe "accuracy" is the most appropriate measure.

\begin{equation}
\label{eqn:acc}
accuracy = \frac{1}{n}\sum_{i}z_i \quad \left\{\begin{matrix}
z_i=1 & if & p_i = c_i \\ 
z_i=0 & if & p_i \neq c_i
\end{matrix}\right.
\end{equation} 

\subsection{RQ1: Accuracy on manually annotated data} 
In this section, we answer the question about the accuracy of our machine learning and deep learning techniques on manually annotated data of 8600  reviews.

\subsubsection{Construction of the models}
\hfill \\
{\bf Model with handcrafted features}: For each of the 8600 reviews in our training set, we extract the features using the rules described in Section~\ref{sec:features}. For the \emph{ReviewLength} and \emph{ReadabilityScore} we transformed the score values into a scale of 0 to 1 as follows.

For \emph{ReviewLength} we automatically got the count of words in all the reviews. We then sort and find the percentile to which a review belongs. If a review is in the $60^{th}$ percentile, the feature's value will be $0.6$ and so on.
%Reviews in the 0 to 20 percentile are assigned an ordinal value of 0., 21-40 a value of 2 and so on.

%For \emph{SentimentScore}, we obtained the sentiment score for each sentence in a review and computed the average sentiment score for a review. We then transformed the score into an ordinal scale of 1 to 5 as described above for the \emph{Review Length} feature.
We obtained the Flesch-Kincaid readability score
~\cite{readability1} for each review text and transformed the scores into a scale of 0 to 1 as above.

%We then built (learnt) a model using the above feature values and 9 different traditional machine learning classifiers such as Decision Trees, Naive Bayes and so on using Weka 3.8~\cite{weka}

We then built 9 different traditional machine learning models mentioned previously and trained them using the extracted feature values with Weka 3.8~\cite{weka}.

{\bf Model with Word Vectors:} We constructed a vocabulary of all the words present in our review corpus by removing stop words and converting the rest into lowercase. We then use 100 dimensional glove vectors~\cite{pennington2014glove} pre-trained on Wikipedia-14 and Gigaword 5 datasets to represent each word in our vocabulary. Finally, these word vectors were used to train the machine learning classifiers mentioned previously.
 
{\bf Model using DCNN:}  For DCNNs, no pre-processing of data is required. Therefore, we only removed all special characters in the review text and used it for training.
  
\subsubsection{Results}

To evaluate all these models, we used ten fold cross validation. Here the 8600 reviews were divided into ten equal sets. In a single iteration, nine sets were used for training and one set was used to test. We had ten such iterations, with each of the ten sets serving as the test data once.

The cross-validation accuracy of the models with different traditional machine learning techniques is shown in Figure~\ref{figs:rq1}. The lightly shaded bars i.e., the bars that appear on the left of each pair of bars represent the cross-validation accuracy of each machine learning classifier using the features described in Section~\ref{sec:features}. The best accuracy of 74.9\% was obtained with the IBk classifier i.e., the Instance Based classifier, followed by the J48 (Decision Tree) classifier which had an accuracy of 73.6\%.

The darkly shaded bars i.e., the bars that appear on the right of each pair of bars represent the cross-validation accuracy of each machine learning classifier using word vectors. The best accuracy of 84.52\% was obtained with the One R classifier, followed by the J48 (Decision Tree) classifier which had an accuracy of 83.22\%.

\textit{The DCNN outperformed all the models shown in Figure~\ref{figs:rq1}. It had an accuracy of 92\%.} 

\subsubsection{Discussion}

The deep learning approach appears to perform the best among the three techniques. The time taken by all three approaches is of the same order with the entire model building (i.e., learning) and ten fold cross validation finishing in a few minutes. The feature engineering based approach did not perform as well due to limitations in identifying all possible features which can accurately help in identifying the rating. DCNN outperformed the word vector based model as it captures additional non-sequential context of a word, which may not have been captured by the word vector model. This additional context appears to help the model to classify the review in a more accurate fashion.

Thus, for answering the next two research questions, we consider only DCNN as the automated solution for predicting review mismatches, as it has performed the best on the manually annotated data set.

\subsection{RQ2: Accuracy on data from the wild} 
%In this section, we answer the question as to what is the accuracy of our technique in automatically predicting the star rating of a given review text. 
{\bf Evaluation with users:} We recruited 23 users within our organization for this evaluation. All of them have advanced degrees in computer science. We asked them to write reviews for 5 mobile apps of their choice and provide a suitable rating corresponding to their reviews. At a high level, we instructed the users not to choose all positive or all negative experience apps to ensure diversity of review-ratings. Thus we have a total of $23 \times 5 = 115$ mobile app reviews with associated ratings in our test set. The users were not compensated in anyway for their work. They were also not told about the intention behind this exercise.

%\begin{table}
%\center
%\begin{tabular}{ |c|c| } 
% \hline
% Rating & Count \\ 
% \hline
% \hline
% 1 & 9 \\
% 2 & 15 \\
% 3 & 34 \\
% 4 & 35 \\
% 5 &  22 \\
% \hline
%\end{tabular}
%\caption{RQ2: Distribution of ratings provided by 23 users across 115 reviews}
%\label{table:rq1-stats}
%\end{table}

\begin{figure*}
%\begin{figure}
\caption{RQ3 : Distribution of number of reviews for which DCNN predicted the ratings as \emph{Match} or \emph{Mismatch} with user ratings across 10 applications}
\label{figs:rq2}
\centering
\includegraphics[width=0.7\textwidth]{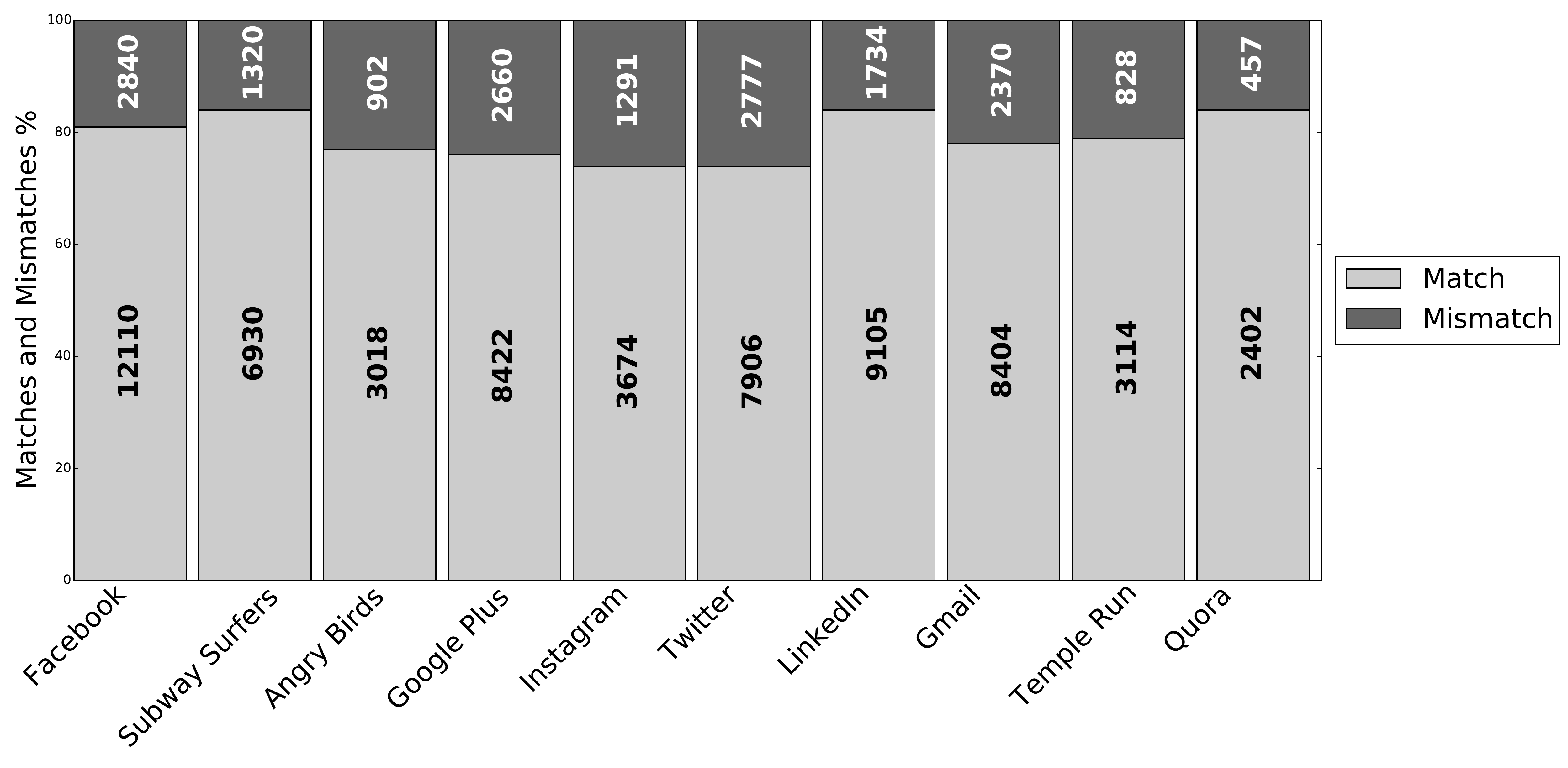}
\vspace{-10pt}
\end{figure*}
%\end{figure}

%Table ~\ref{table:rq1-stats} shows the distribution of the ratings provided by the users.

In total, the 23 different users provided reviews for 66 different apps out of which four apps had figured in our training set viz., Facebook, Twitter, Gmail and LinkedIn. %The diversity in apps is mainly due to the 23 different users along with our high level request to consider apps with which users had a negative experience as well. 
The distribution of 115 reviews among the star ratings were as follows: 9 reviews were rated as one star, 15 as two stars, 34 as three stars, 35 as four stars and 22 as five stars. 

For each review, we then used our DCNN model trained on our manually annotated dataset to predict the rating using \textit{only} the provided review text.
% The deep learning approach was trained on the set of 8600 that the manual annotators had annotated. It was chosen over the other two techniques as it had performed the best on the training set.
%and showed the users the predicted rating along with the rating provided by them. When the predicted rating was different from the user given rating, we asked the users if they felt the
%predicted rating was appropriate and if they were willing to change their rating to
%the predicted rating.

%{\bf Metric:} As discussed before, rating is an inherently subjective task and there is not much difference between a rating of 5 stars and 4 stars, although there is some difference between a rating of 5 stars and 1 stars. Thus, when the predicted rating and the user rating was the same or differed by at most one, we counted the technique's prediction to be correct. Hence, accuracy is given by the number of times the technique's prediction to be correct divided by the total number of reviews in the test set.

{\bf Results:} We computed the accuracy according to equation~\ref{eqn:acc}. We got an accuracy of 87\% if we consider only reviews from those apps which were present in our training set. Whereas, if we consider the entire review set from all the 66 apps, the accuracy was 84\%. We believe that both these accuracy values are fairly good. 
 
%We got an accuracy of 74\% which we believe is a good number as these are reviews drawn from from 66 different apps out of which 62 did not figure in the training set at all. If we consider reviews from only the apps which were there in our training set as well viz., Facebook, Twitter, Gmail and LinkedIn, we get an accuracy of 80\% which suggests that the review rating prediction accuracy for an app can be further improved if the deep learning based approach can be trained on reviews from the same app.

{\bf Discussion:} The accuracy values and the vast number of apps suggest that the DCNN based approach is \textit{fairly good} in assigning an appropriate star rating to a review text and can also be used adequately on apps that have not featured in the training set. The review-rating prediction accuracy for an app can be further improved if the DCNN is trained on review text from the same app. This is due to the fact that sometimes, we see app specific terms in the review text which were not seen by our model previously.

%Note that, one way to evaluate the accuracy of our rating prediction technique is to take random existing reviews for different Android mobile apps on Google Playstore, have users verify whether the provided rating is appropriate and make corrections if deemed to be inappropriate, and finally, compare the user adjusted rating with the rating automatically provided by our technique. However, we decided against the above evaluation methodology as it is both time intensive and also unless the users are very careful they may not be able to correctly judge a review that \emph{they did not write}.

%Another interesting observation from the users was that the users did not consider the rating from an application only perspective. Some users considered reviews as the only way to communicate their needs either in terms of reporting issues to be fixed or enhancement request. They considered rating in a larger context, independent of this review,  such as  \emph{how does the given app perform among all other apps available in that category}. Hence, even though the review might have more ``negative'' connotation to it, they still provided \textit{Good} ratings, as in larger context the app was still the best they had used. 

We also asked the users if they would use a system that would automatically help suggest an appropriate rating for their reviews. %Only five of the 23 reviewers said that they did not need such a system, indicating a need for such an automated solution. Note that, this reinforces our survey results presented in Section~
18 of the 23 users said that they would use such a system, indicating a need for an automated solution. Note that, this reinforces our survey results presented in Section~
\ref{sec:motiv}.

\textbf{Generalizability to other domains: } 
We now discuss about the generalizability of our approach i.e., what would be the accuracy of our approach when applied to non Android app reviews?

We gathered 1000 random product reviews apiece from the publicly available \textit{Amazon MP3} and \textit{Trip Advisor} datasets~\cite{Wang:2011:LAR:2020408.2020505} and applied our DCNN model to obtain star ratings for these reviews. We
then manually verified the results and found that our approach had an accuracy of 88\% on the \textit{Amazon MP3} reviews and 86\% on the \textit{Trip Advisor} reviews. This shows that, our model works well on different domains even without any fine tuning~\cite{fine_tuning} and being trained on a relatively small dataset of an entirely different domain (8600 Android app reviews).  Thus, our model generalizes fairly
well and is not \textit{overfitted} to the training data.
%Among the Amazon MP3 reviews, 350 were classified as being inconsistent with the rating while with the TravelAdvisor reviews, 184 were mismatches. We believe this shows that our approach is fairly generalizable, although our deep learning approach was trained only on a relatively small data set from a completely different domain (mobile apps).

\begin{comment}
\begin{table}
\scriptsize
\centering
\begin{tabular}{ |c|c| } 
 \hline
App & Number of Reviews \\ 
\hline
Angry Birds & 3920 \\ 
Facebook  & 14950 \\ 
Gmail & 10744 \\  
Google Plus  & 11082 \\ 
Instagram & 4965 \\ 
LinkedIn & 10839\\ 
Quora & 2859\\ 
Subway Surfers & 8250 \\ 
Temple Run & 3942 \\ 
Twitter & 10683 \\   
 \hline
 Total & 82234\\
 \hline
\end{tabular}
\caption{Ten Mobile apps and the number of reviews per application considered for identifying mismatch prevalence}
\label{table:dataset-rq3}
\end{table}
\end{comment}

\begin{table}
\scriptsize
\centering
\begin{tabular}{ |c|c||c|c| } 
 \hline
App & Number of Reviews & App & Number of Reviews \\ 
\hline \hline
Angry Birds & 3920 & LinkedIn & 10839\\ 
Facebook  & 14950 & Quora & 2859\\ 
Gmail & 10744 & Subway Surfers & 8250 \\   
Google Plus  & 11082 & Temple Run & 3942 \\ 
Instagram & 4965 & Twitter & 10683 \\   
\hline
\end{tabular}
\caption{Ten Android apps and the reviews per application for identifying mismatch prevalence. 82234 reviews in total}
\label{table:dataset-rq3}
\vspace{-20pt}
\end{table}

\subsection{RQ3: Mismatch Prevalence} 

We now answer the question as to how prevalent are the review-rating mismatches across popular mobile apps for Android on Google Play Store.

{\bf Data:} In order to obtain a count of the review-rating mismatches for an app, we first need to retrieve all the reviews of that app. Unfortunately, Google Play Store does not allow one to download all reviews for an app (unless one is a developer of the app). Thus, the next option is to crawl Google Play Store and retrieve the reviews. Here again, it puts restrictions on crawling and blocks requests if too many of them are sent in a short period of time. Therefore, we decided to retrieve only the reviews from the last few months for each app. 
%We used Scrapy\footnote {\url {http://scrapy.org/}} and Beautiful Soup\footnote {\url {https://www.crummy.com/software/BeautifulSoup/}} to obtain the latest reviews for apps. 
We obtained a total of 82234 reviews of 10 different apps. The apps and the number of retrieved reviews per app are shown in Table~\ref{table:dataset-rq3}. We then ran our DCNN model on these reviews. 

\begin{comment}
	\begin{table*}
		\center
	\begin{tabular}{ |c|c|c|c| } 
	 \hline
	Review & App & Actual & Predicted \\ 
	 &  & Rating & Rating \\ 
	\hline
	\hline
	I love it. I am addicted to it. & &  &\\
	It is a really great game and addictive & Subway Surfers & 1 & 5 \\
	\hline
	App keeps crashing, reinstalled  &  & &  \\
	same thing happens FIX IT ASAP & Facebook & 5 & 1  \\
	\hline
	Love it! New favourite social network! & GooglePlus & 1 & 5 \\
	\hline
	I'm 74 and enjoy it. I can't get friends anymore. &  &  &  \\
	Good game & AngryBirds & 1 & 4 \\
	\hline
	Hi, I cannot get Instagram to update or download,  & & & \\
	just does this download pending screen & & & \\
	for days. I have it set to use my cellular data for download, &  & &  \\
	yet it only says pending & Instagram & 5 & 2 \\
	 \hline
	What I like about quora app is instant valuable feedback & & & \\
	on any query from some of the experienced out there, & & & \\
	not only saves my precious time but help me explore & & & \\
	out element in different dimension in wider array and scope & Quora & 5 & 1\\   
	\hline
	\end{tabular}
	\caption{Sample of mismatched reviews across apps (Automatically identified by our deep learning technique)}
	\label{table:ratings-mismatch-sample-2}
	\end{table*}
\end{comment}

\begin{table}
\scriptsize
\center
\begin{tabular}{ |c|c|c|c| } 
 \hline
Review & App & Actual & Predicted \\ 
 &  & Rating & Rating \\ 
\hline
\hline
I love it. I am addicted to it. & Subway Surfers & 1 & 5 \\
\hline
App keeps crashing, reinstalled  &  & &  \\
same thing happens FIX IT ASAP & Facebook & 5 & 1  \\
\hline
Love it! New favourite social network! & GooglePlus & 1 & 5 \\
\hline
I'm 74 and enjoy it. I can't get &  &  &  \\
friends anymore. Good game & AngryBirds & 1 & 4 \\
\hline
\end{tabular}
\caption{Sample of mismatched reviews across apps (Automatically identified by the DCNN model)}
\label{table:ratings-mismatch-sample-2}
\vspace{-20pt}
\end{table}

{\bf Results:} 
%The same accuracy computation as in equation~\ref{eqn:acc} was done here as well. 
Accuracy was computed as before using equation~\ref{eqn:acc}.
The percentage of reviews for which the original and the predicted star rating categories \textit{matched} and \textit{did not match} are shown in the stacked bar chart in Figure~\ref{figs:rq2}. The results suggest that a substantial number of reviews (17179 or 20\%) have inconsistent star ratings. This phenomenon is not restricted to a few apps but appears across all the apps, ranging from 16\% (for Quora, LinkedIn and Subway Surfers) to 26\% (Instagram and Twitter). A sample of the mismatched reviews are shown in Table~\ref{table:ratings-mismatch-sample-2}.

{\bf Change of average rating due to mismatches:}
%For each app in Table~\ref{table:dataset-rq3}, we computed the average original star rating for the reviews shown in Table~\ref{table:dataset-rq3}. We then computed the average predicted rating using the ratings given by our DCNN model for each app. The original average rating and the predicted average rating after accounting for mismatches is shown in Table~\ref{table:dataset-rq3-new-avg}.

For each app in Table~\ref{table:dataset-rq3}, we computed the average rating, using the original star rating for the reviews shown in Table~\ref{table:dataset-rq3}. We then computed the average rating using the predicted ratings of our DCNN model for each app. The original average rating and the predicted rating after accounting for mismatches is shown in Table~\ref{table:dataset-rq3-new-avg}.

The average rating decreases for some apps but increases for others when we re-calculate after correcting the mismatches.  The average rating variation ranges from 0.3 to 0.7.

\begin{table}
\scriptsize
\centering
\begin{tabular}{ |c|c|c||c|c|c| } 
 \hline
App & Org. Avg. & New Avg.  &  App & Org. Avg. & New Avg.  \\  
 & Rating & Rating &  & Rating & Rating \\ \hline
\hline
Angry Birds & 3.5 & 4.1 & LinkedIn & 3.5 & 3.2\\ 
Facebook  & 2.1 & 1.7 & Quora & 3.8 & 4.2 \\ 
Gmail & 3.2 & 2.6 & Sub. Surf.  & 4.5 & 4.2 \\   
G Plus  & 3.7 & 3.1 & Temple Run & 4.1 & 3.7 \\ 
Instagram & 3.1 & 3.8 & Twitter & 3.4 & 2.8 \\   
\hline
\end{tabular}
\caption{Original and New Average Ratings for ten mobile apps after recomputing rating using DCNN}
\label{table:dataset-rq3-new-avg}
\vspace{-20pt}
\end{table}

{\bf General Discussion:}
Note that, it is possible that such mismatched reviews are not really mismatches but rather mis-predictions by our system. 
%However, \textbf{RQ1} and \textbf{RQ2} suggest that our accuracy is 92\% and 80\% respectively. Thus, we believe that most of the predicted mismatches are actual mismatches. To further validate this, 
%To validate this, we manually analyzed the mismatches by examining 260 random reviews, with 26 apiece from each of the ten apps. Our tool
%had an accuracy of 90\% when considering all the reviews. Of the 260 reviews, there were 88 which were
%classified as mismatch by our system (i.e., 34\%). Among these, we had an accuracy of 71\% i.e., 66 of the 88 were indeed mismatches. Thus, 66 of
%the 260 reviews were mismatches (i.e., 25.4\%)
%To validate this, we manually analyzed the mismatches by examining 510 random reviews, with 51 apiece from each of the ten apps. Our tool
%had an accuracy of 88\% when considering all the reviews. Of the 510 reviews, there were 164 which were
%classified as mismatch by our system (i.e., 32\%). Among these, we had an accuracy of 68\% i.e., 111 of the 164 were indeed mismatches. Thus, overall, 111 of the 510 reviews were indeed mismatches (i.e., 21.8\%)
To validate this, we manually analyzed the mismatches by examining 760 random reviews, with 76 apiece from each of the ten apps. 
%Our tool had an accuracy of 89.6\% when considering all the reviews.
Of the 760 reviews, there were 192 which were classified as mismatch by our system (i.e., 25.26\%). Among these, we had an accuracy of 90.62\% i.e., 174 of the 192 were indeed mismatches. Thus, overall, 174 of the 760 reviews were genuine mismatches (i.e., 23\%)

%We focused on mismatches where the DCNN model predicted the rating as Good (i.e. 5-4) and the actual user provided rating was Bad (1-2) and vice versa (as the subjectivity is reduced in these cases). 

Further, our qualitative observations are as follows: Errors in mismatch prediction happen when reviews contain non-English representations of English words like \textit{``waaaaahoooo''} and \textit{``woooorked!!!''} etc. In apps like \textit{Quora} some reviews talked about the content (questions and answers) served by the app rather than about the app as such. This lead to the detection of a few mismatches incorrectly.

\begin{comment}
\begin{itemize}
\itemsep0em
%  \item Our deep learning based approach predicts mismatches (downgrade %from \textit{Good} to \textit{Bad}) accurately when reviews' context is %about issues or complaints around \textit{``updates''}, %\textit{``advertisements''}, \textit{``freezing or loading''}
%  \item Further in games category (Subway Surfers, Temple Run and Angry %Birds), our approach predicts mismatches (upgrade from \textit{Bad} to %\textit{Good}) accurately, as reviews typically display less variations and usually contain words like \textit{``love''},\textit{``happy''},\textit{``addicted''}.
   \item Errors in mismatch prediction happen when reviews contain non-English representations of English words like \textit{``waaaaahoooo''} and \textit{``woooorked!!!''} etc. 
   \item In apps like \textit{Quora} some reviews talked about the content (questions and answers) served by the app rather than about the app as such. This lead to the detection of a few mismatches incorrectly.
   %So the vocabulary learnt by our deep learning approach did not have a good overlap here and hence detected a few mismatches incorrectly. %(Table~\ref{table:ratings-mismatch-sample-2}, last row) 
\end{itemize}
\end{comment}

\subsection{Replication Package}  Our tool 
%to predict review rating 
is available at \url{http://mismatch.mybluemix.net}. 
The user survey responses, data sets for manual annotation and 
the evaluations are also provided there. Further, model parameters
for the DCNN are also mentioned in detail.

\subsection{Threats to Validity} 

Our study is focused on applications from Android with reviews from Google Play Store, and hence it might not generalize to other distribution platforms like Apple Store. Due to constraints imposed by Google Play Store on downloading all reviews of an app, we had to perforce evaluate on a smaller subset of the latest 82234 reviews from 10 popular Android apps. Thus, our results may not generalize to all reviews, especially when the review text contain app specific terminologies not seen by the model beforehand. To mitigate this we downloaded as many reviews as possible and in future, plan to evaluate our approach on reviews from the Apple Store. 

User surveys are generally prone to various threats\cite{selecting-empirical-methods-for-se-research}
such as being unrepresentative, exhibiting bias and idealistic responses (what people say in a survey can be different from what they actually do in practice). 
To mitigate these threats, we tried to ensure that we obtained responses from a representative sample by posting notifications about the survey on many diverse platforms. We also tried to avoid bias in the responses by not revealing the intentions behind our exercise and avoiding any kind of compensation. Finally, we used the manual annotations to reinforce some of the survey's findings.

In constructing our training set, the rating given by annotators might not be accurate since they are not aware of the circumstances under which the review was written and the original rating provided. To mitigate this we had fairly large sample of 8600 reviews from diverse apps and three independent annotators.

In the user evaluation, there is a possibility for certain users to not rate their reviews in accordance to what is expected. To mitigate this, we conducted the evaluation with a fairly large pool of 23 participants with each reviewing 5 different apps of their choice.

\section{Implications of our work}
\label{sec:impl}
%implications.tex
We believe this is a foundational work and
can be used in several prior research works 
such as~\cite{Chen,Linares}. %Our future work will focus on this aspect of gauging the improvement obtained by identifying review rating mismatches.
Prior research that use reviews and rating, make an assumption that the rating and reviews match and thus can be used as is (i.e, the average rating truly represents the experience of the end-users;
a low star rating implies a negative opinion and so on). 

Our work strongly suggests that we need to be more careful in dealing with reviews and rating. (For ex: a heuristic for automatically
finding negative reviews could assume a low star rating of 1 or 2. However, this heuristic may not be very accurate and would miss a number
of negative reviews which have been erroneously rated 4 or 5 stars). 

Further as shown in Table~\ref{table:dataset-rq3-new-avg}, 
review-rating mismatch will affect the overall average rating of an app and this can impact research that uses
average rating~\cite{Linares} to determine success of an app.

There is a strong correlation between average rating and downloads
~\cite{appStoreMiningMsr2012}. Mismatched review-ratings can deflate average rating leading to fewer downloads and
consequently a loss of revenue.
Small and upcoming apps with less number of downloads and reviews are especially affected by these inconsistent review-ratings. 

Our future work will focus on determining the improvement in existing
research that can be obtained by identifying review rating mismatches.

\section{Related Work}
\label{sec:relatedwork}
%related work

Broadly, most of the related work has focused on analyzing the content of the review text using techniques such as topic modeling to identify bugs and feature requests~\cite{VuMuo,Iacob,Galvis,Chen}.

%Average review rating has been used as a measure of the success of an %app in
%\cite{Linares} where they try to understand the relation between the success of an 
%Android app and change/fault proneness of the underlying Android APIs.

To help developers prioritize the devices to test their app, \cite{Khalid}
examined reviews from different devices for the same app and found that some devices gave significantly lower ratings.

Dave et al. \cite{Dave} use information retrieval techniques to distinguish between positive and negative product reviews. They state that the performance of their method is affected due to \emph{rating inconsistencies}, which they 
define as \emph{``similar qualitative descriptions yield very different quantitative reviews from reviewers. In the most extreme case, reviewers do not understand the rating system and give a 1 instead of a 5"}. Our work in contrast addresses the rating inconsistency problem directly.

Fu et al. \cite{FuBin} try to understand why people might dislike an app. As part of this, they allude to the presence of reviews with inconsistent ratings. They propose a simple approach of using regression on words with a certain frequency (viz., $>$ 10).
They unfortunately do not provide an accuracy assessment of their approach. The first of our three approaches can be seen as a generalization of their approach as it uses TF-IDF and further allows the use of different classifiers apart from regression. Also, our DCNN approach performs better than the mentioned approach.

\section{Conclusion}
\label{sec:conc}
%conclusion.tex
\begin{comment}
Mobile app distribution platforms such as Google Play Store or Apple Store
allow users to share their feedback about the apps downloaded and used by 
them. Users share their feedback via a one to five star rating along with 
a corresponding review text. However, due to a variety of reasons, the
rating provided might not be consistent with the review text. For example,
a review such as ``Great App!" might be accompanied by an (inconsistent) one star rating. Such inconsistent ratings bring down the average rating of an app which affects the app downloads as studies suggest that users are highly influenced by
average rating of an app when making a decision about downloading the app. Small and upcoming apps with less number of downloads and reviews especially are affected by these inconsistent review-ratings. 
\end{comment}

In this paper, we performed a study of the Android app review-rating mismatch problem. We conducted a survey of Android app end users and developers. The survey responses suggest that: (1) review text and corresponding star ratings should match; (2) it is useful to have an automated system to detect mismatches; (3) end users do not update the star ratings when they update their review text; (4) developers believe mismatches are prevalent and affects overall app development.

We manually analyzed 8600 reviews from 10 mobile apps available for Android. These apps include Facebook, Gmail and other popular apps. We found that about 20\% of the reviews had ratings which did not
match with the review text. Further, our study suggested that manually analyzing reviews to detect inconsistent ratings can be tedious and time consuming, thus, warranting automated solutions.

We developed multiple automated systems to detect reviews with inconsistent ratings. These systems are based on machine and deep learning methods. We then empirically established that our Dependency based Convolutional Neural Network model performs well in practice i.e., can accurately identify reviews whose rating does not match with the opinion expressed in the review text. Our system achieved an accuracy of 92\% on the manually annotated data.

Further, we performed an end user evaluation. We recruited 23 Android app end-users and asked them to write reviews for any five mobile apps used by them, along with providing a rating ranging from one to five stars. We predicted the star rating for these user reviews using DCNN and compared with the user provided rating. Our system achieved an upward accuracy of 87\%.

We finally used our system to detect review-rating mismatches across 10 popular apps on Android (available on Google Play Store) and found that mismatched review-ratings are fairly prevalent across apps ranging from 16\% to 26\%.

%\newpage

\bibliographystyle{ACM-Reference-Format}
\bibliography{references}  
\end{document}